\documentclass[
]{ceurart}

\usepackage{listings}
\usepackage{todonotes}
\begin{document}

\copyrightyear{2026}
\copyrightclause{Copyright for this paper by its authors.
  Use permitted under Creative Commons License Attribution 4.0
  International (CC BY 4.0).}

\conference{Joint Proceedings of the ACM Intelligent User Interfaces (IUI) Workshops 2026, March 23-26, 2026, Paphos, Cyprus}

\title{Agentic AI for Safety-critical Multi-drone Systems: Challenges and Opportunities}



\author[1]{Timothy Merritt}[%
orcid=0000-0002-7851-7339,
email=merritt@cs.aau.dk,
url=https://www.ixd.net,
]
\cormark[1]
  
\author[2]{Alejandro Jarabo-Peñas}[%
orcid=0000-0002-3312-7062,
email=alejp@mmmi.sdu.dk,
]

\author[2]{Juan Bravo-Arrabal}[%
orcid=0000-0002-4046-1528,
email=juanba@mmmi.sdu.dk,
]

\author[1]{Maria-Theresa Bahodi}[%
orcid=0000-0003-4512-400X,
email=mtoh@cs.aau.dk,
]

\author[2]{Anders Lyhne Christensen}[%
orcid=0000-0002-9994-2908,
email=andc@mmmi.sdu.dk,
url=https://anderslyhnechristensen.com/,
]

\address[1]{Human-centered Computing (Aalborg University),
  300 Selma Lagerløfs Vej, Aalborg Øst, 9220, Denmark}
  \address[2]{The Maersk Mc-Kinney Moller Institute (University of Southern Denmark),
  Campusvej 55, 5230, Odense, Denmark}

\cortext[1]{Corresponding author.}


\begin{abstract}
Multi-drone systems are increasingly positioned for safety-critical missions such as search and rescue (SAR) and critical infrastructure monitoring. Yet, real-world adoption remains constrained not only by autonomy performance, but by the difficulty of integrating agentic behavior into professional work: operators must understand, trust, and govern automation under uncertainty, time pressure, and accountability. This position paper synthesizes the ambitions and lessons from two ongoing efforts: NAMUR, which explores LLM-supported robot control in SAR and firefighting contexts, and PERSIST, which explores persistent drone operations for monitoring and security at critical infrastructure sites. We argue that agentic AI should be approached as a socio-technical design problem, where interfaces, oversight mechanisms, and evaluation practices are as critical as algorithms. We outline a human-centered, participatory, and iterative research approach aimed at uncovering stakeholder needs, shaping agent capabilities through successive prototypes, and producing transferable proof-of-concept systems and evaluation strategies for other safety-critical contexts.
\end{abstract}


\maketitle

\section{Introduction}
Multi-robot and multi-drone systems promise new capabilities for safety-critical work, including emergency response~\cite{Hoang2023} and security and inspection at critical infrastructure~\cite{jacobsen2023design}. At the same time, these domains impose strict demands~\cite{murphy2008search,ventura2012search,drew2021multi}: operations are uncertain, accountability is high, and workflows are governed by professional roles, protocols, and risk management. In such settings, it is rarely sufficient for autonomy to ``work'' in a technical sense; it must also be understandable, reliable, governable, and adoptable.

Controlling multiple robots as a \emph{single operational system} amplifies these demands~\cite{christensen2022herd,bahodi2025before,chung2018survey}. As fleet size grows, the traditional one-operator/one-vehicle model breaks down~\cite{planke2020cyber}: coordination overhead increases, maintaining shared situation awareness becomes harder, and small uncertainties can cascade into safety issues (e.g., conflicting task priorities, deconfliction problems, or ambiguous responsibility for who approved what). Operators therefore need to interact at multiple levels of granularity~\cite{kim2020user}. That might entail setting mission-level intent and constraints for the group, while still being able to inspect, redirect, or take control of a specific drone or subset when conditions change.

Recent progress in agentic AI and large language models (LLMs) has renewed interest in natural-language tasking, mixed-initiative planning, and autonomous execution in robotics~\cite{wang2024large}. However, these capabilities introduce new interaction challenges~\cite{sapkota2025uavs}: how should agent intentions be represented; how do operators supervise, approve, and override decisions; how can correct behavior be guaranteed in the messy realities of critical field operations?

This paper is organized as follows. Section~\ref{sec:cases} introduces the two motivating project contexts---NAMUR (SAR and firefighting) and PERSIST (persistent critical-infrastructure monitoring)---and distills shared interaction requirements for multi-drone control. Section~\ref{sec:agentic_agenda} synthesizes lessons from our prior field engagements to frame the key opportunities and safety-critical tensions that motivate \emph{governable} agentic autonomy, and outlines our participatory, iterative approach for shaping such systems with stakeholders. Building on these requirements, Section~\ref{sec:architecture} presents an LLM-based multi-agent system (LLM-MAS) that decomposes natural-language intent into reviewable sub-tasks, constrains execution through deterministic tools, and enforces operator preview and approval. Finally, Section~\ref{sec:conclusions} concludes with implications for future agentic multi-drone systems in safety-critical work.

\section{Case Contexts and Project Objectives}
\label{sec:cases}

\begin{figure}[t]
    \centering
    \includegraphics[width=.95\linewidth]{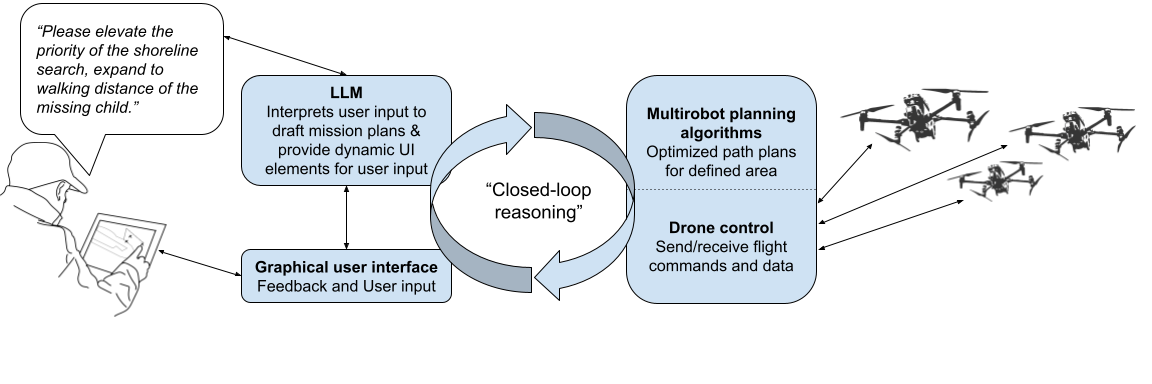}
    \caption{NAMUR concept illustration: an LLM-driven natural language interaction system with ``closed-loop reasoning'' to support human tasking and robot control in emergency response.}
    \label{fig:namur_overview}
\end{figure}

Our position is shaped by two complementary projects that rely on agentic AI in different safety-critical conditions: (i) time-critical emergency response (NAMUR) and (ii) long-horizon, persistent operations at critical infrastructure (PERSIST). We summarize each context in terms of operational setting, stakeholders, objectives, and the agentic-AI questions that emerge.

\subsection{NAMUR: Agentic support for SAR and firefighting}
\label{subsec:namur}
NAMUR investigates how LLM-supported, agentic interaction can help teams coordinate a multi-drone system during emergency response work such as search and rescue and firefighting support (see Figure~\ref{fig:namur_overview}). The operational setting is characterized by time pressure, dynamic hazards, incomplete and rapidly changing situational information, and a strong need for coordination across roles~\cite{Hoang2023,delmerico2019current}. Deploying multi-robot systems in the field also depends on robust communication and edge-cloud infrastructure~\cite{bravo2021internet}, and requires interfaces that scale across heterogeneous robot platforms. Because operators with different roles and expertise tend to interact differently with each robot type, agentic AI can serve as a unifying abstraction layer---providing a more coherent, comprehensible relationship between users and diverse physical systems. In these contexts, technology is only useful when it integrates with established command structures and communication practices, and when responsibility for high-stakes decisions remains clear.

\paragraph{Stakeholders and work setting.}
The primary stakeholders include incident commanders and coordinators, field responders operating in hazardous environments, and technology operators responsible for robotic assets. The work is organized through role-specialized decision-making and structured communication, where updates and tasking must be concise, timely, and auditable~\cite{jensen2016incident,wolbers2013common}. This setting foregrounds interaction demands that go beyond natural language control: operators must be able to supervise, confirm, and constrain what an agentic system does under uncertainty.

\paragraph{Project objectives.}
NAMUR's objectives are to:
(1) enable higher-level tasking of robotic assets (e.g., translating intent into feasible robot actions);
(2) support mixed-initiative interaction where the system proposes actions, but humans authorize execution;
(3) make agentic behavior interpretable enough for operators to assess risk, timing, and consequences; and
(4) study these interactions in realistic, mission-oriented exercises to surface failure modes, governance needs, and usability constraints that would be invisible in purely lab-based evaluations.

\paragraph{Why agentic AI is compelling here.}
Agentic AI can help transform fragmented inputs (radio updates, map cues, observations) into actionable suggestions and structured plans~\cite{goecks2023disasterresponsegpt, sadik2025humanllm}, and can reduce coordination overhead by maintaining continuity across rapid task switches. At the same time, emergency response makes the limits of agentic AI especially visible: uncertainty is unavoidable, and behavior designed for convenience can become harmful without explicit authorization workflows~\cite{robey2024jailbreaking, cleland-huang2025guardrails}, conservative defaults, and clear escalation boundaries.

\subsection{PERSIST: Persistent multi-drone operations for critical infrastructure}
\label{subsec:persist}
\begin{figure}[t]
    \centering
    \includegraphics[width=.95\linewidth]{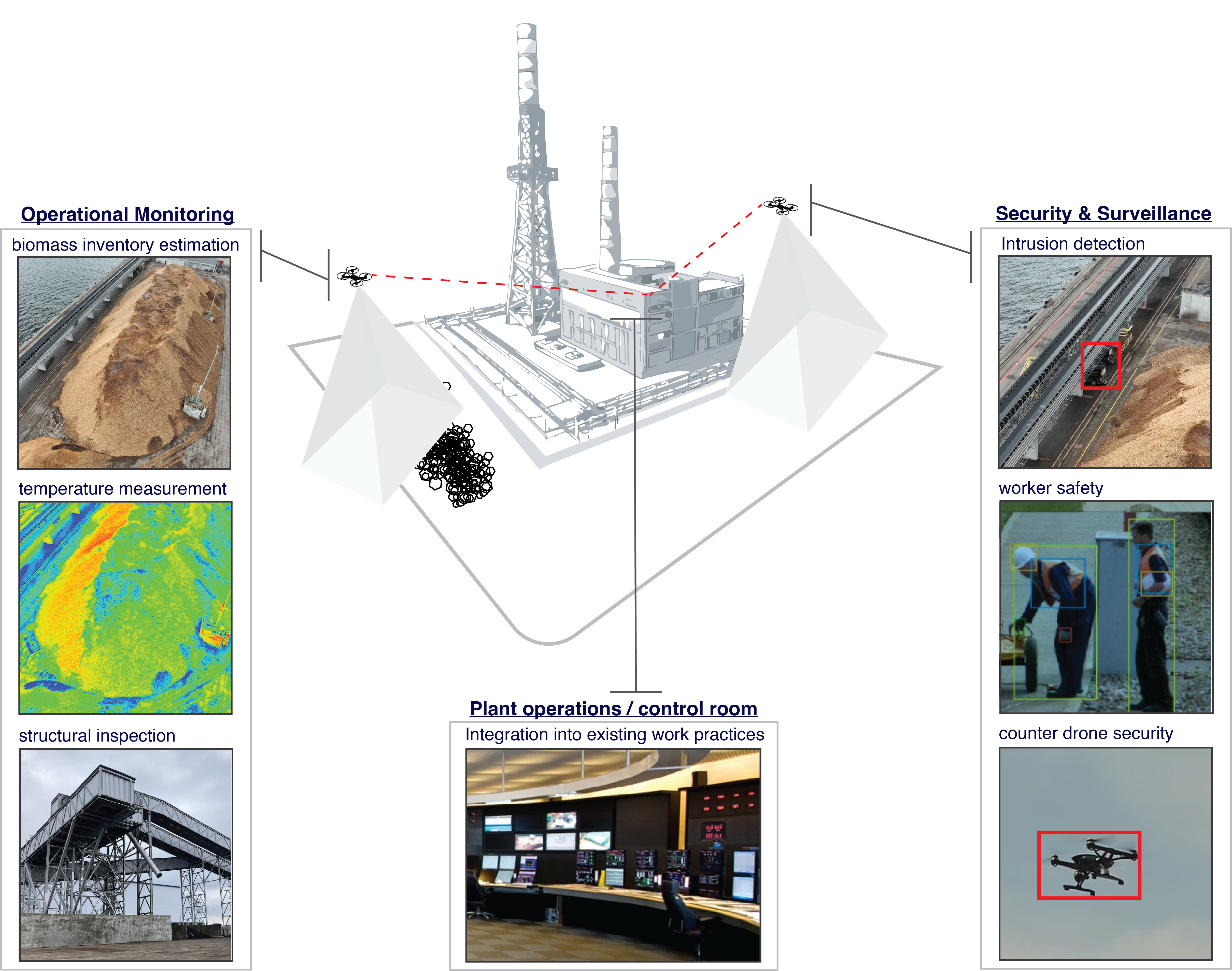}
    \caption{PERSIST overview illustration (from proposal material): examples of task types motivating persistent multi-drone operations (e.g., routine monitoring, inspection, and security workflows).}
    \label{fig:persist_tasks}
\end{figure}

PERSIST investigates persistent, multi-drone operations for monitoring, inspection, and security at critical infrastructure sites (see Figure~\ref{fig:persist_tasks}). In contrast to emergency response, the dominant challenge here is not a single high-tempo mission but sustained operations across days and shifts, where drones must repeatedly execute routine tasks, respond to anomalies, and integrate into existing organizational workflows for security and maintenance. Recent work on large language model (LLM) agents in industrial automation highlights the broader relevance of agentic approaches for orchestrating complex operational systems beyond the lab~\cite{xia2023llm_agents_industry}.

\paragraph{Stakeholders and work setting.}
Key stakeholders include site security personnel, operations and maintenance staff, and organizational decision-makers responsible for safety, compliance, and continuity of service. The setting includes constrained physical environments (restricted zones, sensitive assets), operational rhythms (shift handovers, scheduled inspections), and an expectation that systems behave predictably and recover gracefully from routine disruptions (weather, connectivity, access changes, false alarms).

\paragraph{Project objectives.}
PERSIST's objectives are to:
(1) move from ad hoc drone flights to persistent, repeatable operations that can be delegated and scheduled;
(2) reduce the barrier to entry for professional drone use through interface support for planning, monitoring, and handover;
(3) enable scalable supervision where one operator can manage multiple vehicles and mission types; and
(4) prototype and validate the approach in realistic infrastructure settings, producing proofs of concept with transfer potential to other critical sites.

\paragraph{Why agentic AI is compelling here.}
Agentic AI can support long-horizon orchestration: selecting and parameterizing routine missions, adapting schedules as conditions change, and assisting with anomaly triage and reporting. The risk profile differs from NAMUR: failures may be less acute in the moment, but persistent operations amplify the importance of reliability, drift management, operator fatigue, and organizational trust. Here, the central interaction challenge is sustained governability: users must be able to understand what the system has been doing over time, why exceptions occurred, and what remains unresolved at shift handover.

\subsection{Cross-cutting themes and contrasts}
\label{subsec:crosscut}
Together, NAMUR and PERSIST highlight complementary stressors for agentic multi-drone systems.

\textbf{Tempo and horizon.} NAMUR emphasizes rapid decision cycles and short operational windows; PERSIST emphasizes persistent activity, repeatability, and organizational integration across shifts.

\textbf{Governance and accountability.}
Both contexts demand clear authority and auditability, but in different forms: emergency response requires immediate confirmation and escalation pathways, while infrastructure operations require policy-aligned routines, handover summaries, and post hoc accountability.

\textbf{Implication for our research approach.}
Across both cases, we treat agentic AI as a socio-technical design material: we study how people coordinate, decide, and verify; we prototype agentic behaviors that remain governable; and we iteratively shape the system with stakeholders rather than presuming a final autonomy solution.

\begin{figure}[t]
    \centering
    \includegraphics[width=\linewidth]{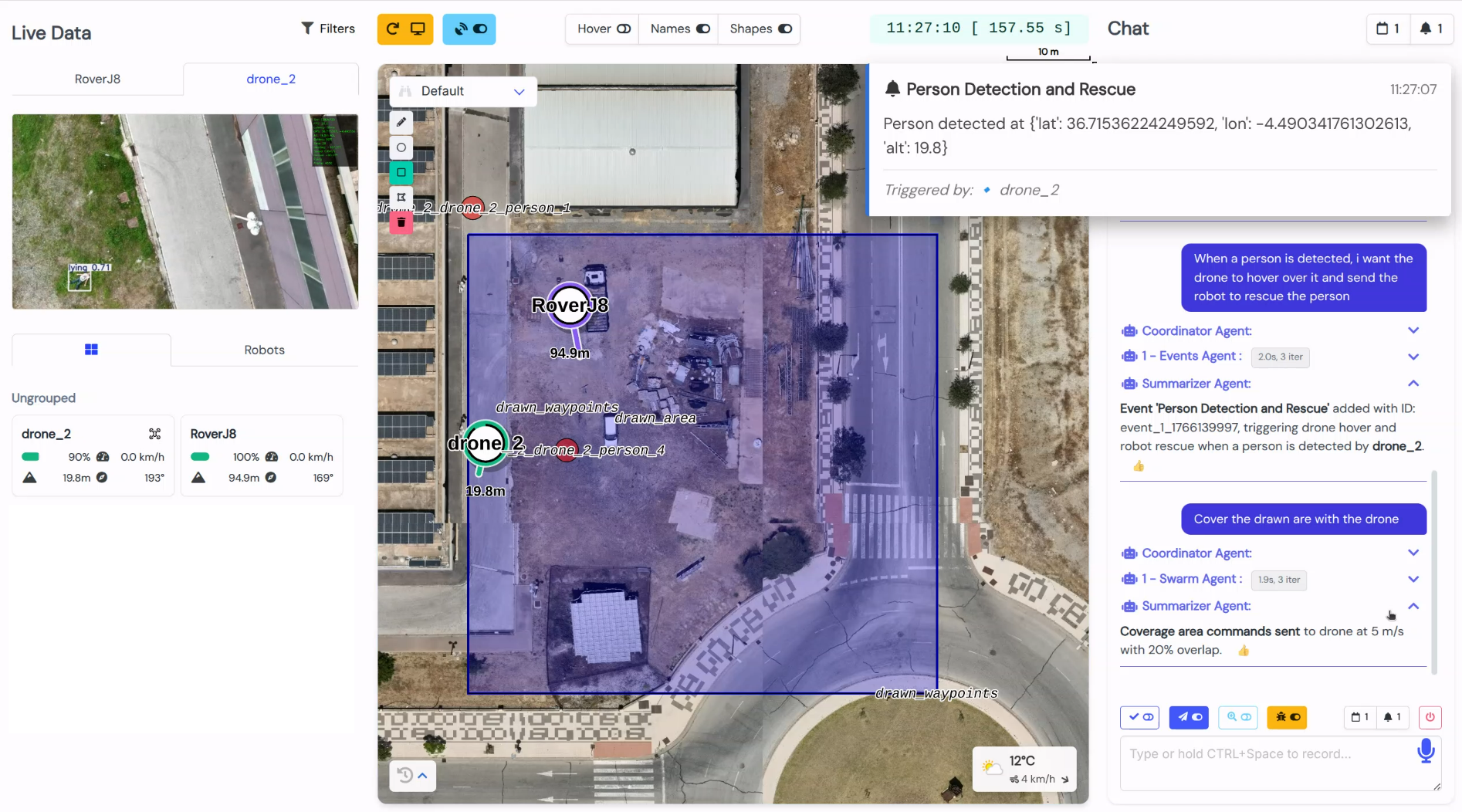}
    \caption{Latest prototype interface screenshot: used in participatory walkthroughs to elicit requirements for oversight, uncertainty communication, and operational fit.}
    \label{fig:prototype_ui}
\end{figure}

\section{Agentic AI for Multi-Drone Operations in Safety-Critical Work: Lessons, Tensions, and an Agenda}
\label{sec:agentic_agenda}
Our perspective on agentic AI is grounded in prior human-centered studies of multi-robot systems in safety-critical environments, including co-design and prototype evaluations with Danish and Spanish emergency services in search-and-rescue (SAR) and firefighting contexts~\cite{Hoang2023,jarabo-penas_real-world_2025}, as well as co-design work with security personnel at a power plant. Across these contexts, practitioners want autonomy that is \emph{self-sustaining} enough to be operationally viable, yet \emph{transparent}, easily learnable, and aligned with local protocols and expertise. This motivates a cautiously optimistic stance: agentic AI can reduce the coordination burden and accelerate sensemaking, but only if paired with interaction mechanisms that keep humans able to authorize, supervise, and diagnose actions under uncertainty. In the ideal scenario, a single operator can supervise a mission involving a fleet of drones through mixed-initiative control: high autonomy for routine progress and rapid intervention at multiple levels of granularity when risk or uncertainty increases.

\subsection{Opportunities that motivate agent support}
Agentic AI is compelling in these domains not because it can replace professional judgment, but because it can help teams manage \emph{scale} (multiple vehicles, multiple data streams, multiple stakeholders) without collapsing situation awareness~\cite{endsley_toward_1995}. Across our cases, the most salient opportunities are: (i) selective attention and summarization that reduces continuous manual scanning, (ii) mixed-initiative planning and replanning that maintain coherent coverage as information changes, and (iii) long-horizon orchestration for persistent operations, including scheduling, anomaly triage, and reporting aligned to staffing constraints.

\subsection{Safety-critical tensions that require governability}
The same capabilities introduce tensions that cannot be addressed by autonomy alone. First, practitioners need systems that can operate without constant attention, yet resist black-box behavior that cannot be explained, constrained, or overridden. Second, strong cueing and alerting can improve short-term performance but encourage attention tunneling, reducing holistic overview during supervision~\cite{bahodi_show_2024}. Third, trust must be calibrated under imperfect perception and dynamic conditions~\cite{ahlskog2024}; conservative defaults, uncertainty cues, and verification workflows are essential. Finally, adoption depends on operational fit: alignment with roles, protocols, and accountability practices, including auditable authorization and decision trails.

\subsection{Research approach: participatory shaping of agentic autonomy}
Rather than aiming for a single, final autonomy concept, we treat agentic AI as a socio-technical design material that must be iteratively shaped through stakeholder engagement with functional prototypes~\cite{bodker-et-2018}. 

We begin with needs discovery grounded in concrete scenarios and demonstrations to map work practices, decision points, and cognitive bottlenecks, then use successive prototypes (see \autoref{fig:prototype_ui} for an example of our functional multi-drone control interface prototype deployed in a real-world context~\cite{jarabo-penas_real-world_2025} \& companion video of live control of UGVs with LLM-enabled voice commands.\footnote{Video showing live control of UGVs with LLM-enabled voice commands:~\url{https://www.youtube.com/watch?v=Hna7AtDS5wo}}) as boundary objects to negotiate what the agent should do, what must require explicit authorization, and what evidence and uncertainty cues the system must surface.

Across iterations, this process yields actionable interaction requirements for \emph{governability}: configurable constraints that domain experts can tune quickly, clear intervention points for high-stakes actions, and observability features that let both operators and developers diagnose behavior and learn from incidents.

\paragraph{Bridge to architecture.}
Together, these cases and lessons motivate an architecture that decomposes intent into reviewable sub-tasks, constrains action through deterministic tools, supports scheduling and long-horizon memory, and enforces explicit operator preview and approval before execution.

\section{An LLM-MAS Architecture for Multi-drone Control}
\label{sec:architecture}
\begin{figure}[tb]
\centering
\includegraphics[width=8cm]{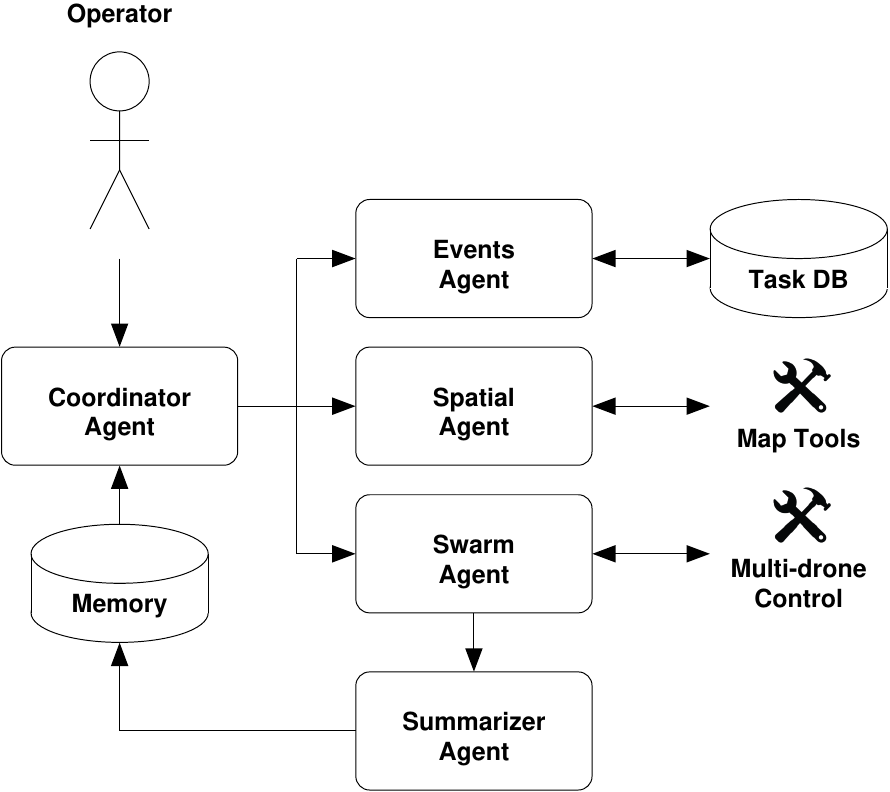}
\caption{Architecture of the LLM-MAS use in the functional prototype.}
\label{fig:llm-arch}
\end{figure}

The functional prototype implements a unified agentic AI architecture that meets the requirements of the NAMUR and PERSIST projects and has been demonstrated in live control of UGVs and UAVs~\cite{jarabo-penas_real-world_2025}. The architecture is implemented as a LLM-MAS composed of five specialized agents: \emph{Coordinator}, \emph{Event}, \emph{Spatial}, \emph{Swarm}, and \emph{Summarizer}. Several agents, including the Spatial Agent and the Swarm Agent, interface with external software tools that provide deterministic capabilities such as spatial computation, multi-drone motion execution, memory storage, and task scheduling. Collectively, these components enable the decomposition of high-level natural language commands into well-defined sub-tasks~\cite{singh2023progprompt,tsushima2025task} that can be previewed, validated, and explicitly approved by an operator prior to execution (Figure~\ref{fig:llm-arch}).

The specific set of tools available to the LLM-MAS is deployment-dependent. For SAR scenarios, dedicated tools support functionalities such as compliant coverage path planning. In contrast, for persistent drone operations at biomass power plants, specialized tools are provided for generating flight paths and image acquisition plans that enable accurate biomass estimation.

In the following, we describe the responsibilities of each agent using a representative operator command as an example.

\begin{itemize}
\item \textbf{Coordinator Agent:} Serves as the central orchestrator of the system. It receives the operator’s command, infers the underlying intent, and decomposes it into concrete sub-queries and execution requests that are delegated to subordinate agents.

\item \textbf{Events Agent:} Manages events and periodic, delayed, and future-oriented directives issued by the operator. It maintains and monitors scheduled and recurring tasks (e.g., \textit{``Estimate biomass volume every weekday at 2 pm''} or \textit{``Return all drones to base at 18:00''}) and ensures that they are executed or surfaced at the appropriate time and context.

\item \textbf{Spatial Agent:} Provides spatial reasoning and situational awareness by accessing a GeoJSON-formatted map alongside the live positions of all active drones and tracked first responders. It identifies relevant semantic map features (e.g., the \textit{``eastern tree line''} or the \textit{``biomass pile next to the water''}) and accesses tools to compute spatial relationships between entities as required.

\item \textbf{Swarm Agent:} Handles motion planning and task execution requests for one or more available drones. It selects suitable drone platforms based on availability and operational context and issues coordinated control commands when multiple drones are involved~\cite{grontved2023decentralized}. Given the real-world consequences of these actions, all requests generated by this agent are subject to operator preview and confirmation. Upon receiving target coordinates from the Spatial Agent, the Swarm Agent invokes an execution tool that publishes the desired actions together with the selected drone identifier(s) to a ROS topic, thereby triggering the downstream planning and flight control pipeline.

\item \textbf{Summarizer Agent:} Generates a concise, human-readable summary of the actions executed and the resulting system state after each command. In addition to being presented to the operator as confirmation, these summaries are persistently stored in the system’s interaction memory. The stored summaries provide a compact, structured record of prior interactions and outcomes, which is made available as contextual input to the Coordinator Agent in subsequent dialogue turns. This persistent context enables coherent multi-turn interactions and reference resolution across commands, such as interpreting follow-up instructions that rely on previously mentioned entities or actions (e.g., \textit{``Move drone 5 to the main building''} followed by \textit{``Now, move it to the nearest safe landing zone''}).
\end{itemize}

By explicitly separating responsibilities across specialized agents and constraining their behavior through structured prompts and tool interfaces, the proposed LLM-MAS enables transparent, interpretable, and safe translation of natural language instructions into coordinated multi-drone mission execution. Furthermore, the integration of persistent interaction history and explicit task management supports robust multi-turn dialogue and long-horizon autonomy under continuous operator supervision.

\section{Conclusion}
\label{sec:conclusions}
Agentic AI holds promise for scaling multi-drone operations in SAR and critical infrastructure monitoring, but introduces challenges around oversight, trust calibration, operational fit, and evaluation. Grounded in NAMUR and PERSIST, we framed these requirements and presented an LLM-MAS architecture that operationalizes governable agentic autonomy through constrained tool use and explicit operator preview/approval. This framing supports future research in autonomy and provides interactive infrastructure required to deploy agentic systems responsibly.

Looking ahead, we see three implications for future agentic multi-drone systems in safety-critical work. First, \textbf{governability should be treated as a core system property}, not a UI add-on: architectures should make authorization points, constraints, and intervention mechanisms explicit and auditable. Second, \textbf{agentic behavior should be built from constrained, inspectable tool use} rather than unconstrained free-form action, enabling predictable failure modes, reproducible debugging, and clearer responsibility boundaries. Third, \textbf{evaluation should reflect real operational use}. This means measuring more than task success: we should assess situation awareness (and whether alerts cause attention tunneling), trust calibration under uncertainty, coordination overhead, and whether the system supports accountable after-action review.

\begin{acknowledgments}
This work was supported by the Innovation Fund Denmark for DIREC project U07 (the PERSIST project), and the Independent Research Fund Denmark under grant 10.46540/4264-00105B (the NAMUR project).  
\end{acknowledgments}

\section*{Declaration on Generative AI}
 During the preparation of this work, the authors used Grammarly and GPT 5.2 in order to: Grammar and spelling check. After using these tools, the authors reviewed and edited the content as needed and take full responsibility for the publication’s content.

\bibliography{references}

@inproceedings{ahlskog2024,
author = {Ahlskog, Johanna and Bahodi, Maria-Theresa and Lugmayr, Artur and Merritt, Timothy},
title = {Fostering Trust Through User Interface Design in Multi-Drone Search and Rescue},
year = {2024},
isbn = {9798400709890},
publisher = {Association for Computing Machinery},
address = {New York, NY, USA},
url = {https://doi.org/10.1145/3686038.3686052},
doi = {10.1145/3686038.3686052},
booktitle = {Proceedings of the Second International Symposium on Trustworthy Autonomous Systems},
articleno = {2},
numpages = {11},
location = {Austin, TX, USA},
series = {TAS '24}
}

@inproceedings{jarabo-penas_real-world_2025,
  title={Real-{World} {Deployment} of an {LLM}-{Enabled} {Voice}-{Commanded} {UGV} for {Logistics} in {SAR} {Missions}},
  author={Jarabo-Pe{\~n}as, Alejandro and Bravo-Arrabal, Juan and Lin-Yang, Da-hui and Pastor, Francisco and Ladig, Robert and Fern{\'a}ndez-Lozano, Juan Jes{\'u}s and Christensen, Anders and Garc{\'\i}a-Cerezo, Alfonso},
  booktitle={2025 IEEE International Symposium on Safety Security Rescue Robotics (SSRR)},
  year={2025},
  organization={IEEE}
}

@incollection{murphy2008search,
  author       = {Robin R. Murphy and
                  Satoshi Tadokoro and
                  Daniele Nardi and
                  Adam Jacoff and
                  Paolo Fiorini and
                  Howie Choset and
                  Aydan M. Erkmen},
  bibsource    = {dblp computer science bibliography, https://dblp.org},
  booktitle    = {Springer Handbook of Robotics},
  doi          = {10.1007/978-3-540-30301-5\_51},
  editor       = {Bruno Siciliano and
                  Oussama Khatib},
  pages        = {1151--1173},
  publisher    = {Springer},
  title        = {{S}earch and {R}escue {R}obotics},

  year         = {2008}
}

@inproceedings{ventura2012search,
  author       = {Ventura, Rodrigo and Lima, Pedro U},
  booktitle    = {2012 Third International Conference on Emerging Security Technologies},
  organization = {IEEE},
  pages        = {12--19},
  title        = {{S}earch and rescue robots: {T}he civil protection teams of the future},
  year         = {2012}
}

@article{drew2021multi,
  author       = {Drew, Daniel S},
  doi          = {10.1007/s43154-021-00048-3},
  journal      = {Current Robotics Reports},
  pages        = {189--200},
  publisher    = {Springer},
  title        = {{M}ulti-agent systems for search and rescue applications},
  volume       = {2},
  year         = {2021}
}

@article{jensen2016incident,
  author       = {Jensen, Jessica and Thompson, Steven},
  journal      = {Disasters},
  number       = {1},
  pages        = {158--182},
  publisher    = {Wiley Online Library},
  title        = {{T}he incident command system: a literature review},
  volume       = {40},
  year         = {2016}
}

@article{wolbers2013common,
  author       = {Wolbers, Jeroen and Boersma, Kees},
  journal      = {Journal of Contingencies and Crisis management},
  number       = {4},
  pages        = {186--199},
  publisher    = {Wiley Online Library},
  title        = {{T}he common operational picture as collective sensemaking},
  volume       = {21},
  year         = {2013}
}

@inproceedings{goecks2023disasterresponsegpt,
  author       = {Vinicius G. Goecks and Nicholas R. Waytowich},
  title        = {{DisasterResponseGPT}: {L}arge {L}anguage {M}odels for {A}ccelerated {P}lan of {A}ction {D}evelopment in {D}isaster {R}esponse {S}cenarios},
  booktitle    = {Workshop on Challenges in Deployable Generative AI at International Conference on Machine Learning (ICML)},
  year         = {2023},
  eprint       = {2306.17271},
  eprinttype   = {arXiv},
  doi          = {10.48550/arXiv.2306.17271}
}

@article{cleland-huang2025guardrails,
  author       = {Jane Cleland-Huang and Pedro Antonio Alarcon Granadeno and Arturo Miguel Russell Bernal and Demetrius Hernandez and Michael Murphy and Maureen Petterson and Walter Scheirer},
  title        = {{C}ognitive {G}uardrails for {O}pen-{W}orld {D}ecision {M}aking in {A}utonomous {D}rone {S}warms},
  journal      = {CoRR},
  volume       = {abs/2505.23576},
  year         = {2025},
  eprint       = {2505.23576},
  eprinttype   = {arXiv},
  doi          = {10.48550/arXiv.2505.23576}
}

@article{sadik2025humanllm,
  author       = {Ahmed R. Sadik and Muhammad Ashfaq and Niko M{\"{a}}kitalo and Tommi Mikkonen},
  title        = {{H}uman-{LLM} {S}ynergy in {C}ontext-{A}ware {A}daptive {A}rchitecture for {S}calable {D}rone {S}warm {O}peration},
  journal      = {CoRR},
  volume       = {abs/2509.05355},
  year         = {2025},
  eprint       = {2509.05355},
  eprinttype   = {arXiv},
  doi          = {10.48550/arXiv.2509.05355}
}

@inbook{Hoang2023,
  address      = {Cham},
  author       = {Hoang, Maria-Theresa Oanh
                  and Gr{\o}ntved, Kasper Andreas R{\o}mer
                  and van Berkel, Niels
                  and Skov, Mikael B.
                  and Christensen, Anders Lyhne
                  and Merritt, Timothy},
  booktitle    = {Cultural Robotics:
                  Social Robots and Their Emergent Cultural Ecologies},
  doi          = {10.1007/978-3-031-28138-9_11},
  editor       = {Dunstan, Belinda J.
                  and Koh, Jeffrey T. K. V.
                  and Turnbull Tillman, Deborah
                  and Brown, Scott Andrew},
  isbn         = {978-3-031-28138-9},
  pages        = {163--176},
  publisher    = {Springer International Publishing},
  title        = {{D}rone {S}warms to {S}upport {S}earch and {R}escue {O}perations: {O}pportunities and {C}hallenges},

  year         = {2023}
}

@article{wang2024large,
  title={{Large language models for robotics: Opportunities, challenges, and perspectives}},
  author={Wang, Jiaqi and Shi, Enze and Hu, Huawen and Ma, Chong and Liu, Yiheng and Wang, Xuhui and Yao, Yincheng and Liu, Xuan and Ge, Bao and Zhang, Shu},
  journal={J. Autom. and Intell.},
  year={2025},
  volume ={4},
  issue={1},
  pages={52--64},
  publisher={Elsevier}
}

@article{robey2024jailbreaking,
  author       = {Alexander Robey and
                  Zachary Ravichandran and
                  Vijay Kumar and
                  Hamed Hassani and
                  George J. Pappas},
  bibsource    = {dblp computer science bibliography, https://dblp.org},
  doi          = {10.48550/ARXIV.2410.13691},
  eprint       = {2410.13691},
  eprinttype   = {arXiv},
  journal      = {CoRR},
  title        = {{J}ailbreaking {L}{L}{M}-Controlled {R}obots},

  volume       = {abs/2410.13691},
  year         = {2024}
}

@article{tsushima2025task,
  author       = {Yosuke Tsushima and
                  Shu Yamamoto and
                  Ankit A. Ravankar and
                  Jose Victorio Salazar Luces and
                  Yasuhisa Hirata},
  bibsource    = {dblp computer science bibliography, https://dblp.org},
  doi          = {10.1109/LRA.2025.3531153},
  journal      = {{IEEE} Robotics Autom. Lett.},
  number       = {3},
  pages        = {2383--2390},
  title        = {{T}ask {P}lanning for a {F}actory {R}obot {U}sing {L}arge {L}anguage {M}odel},

  volume       = {10},
  year         = {2025}
}

@article{delmerico2019current,
  author       = {Jeffrey A. Delmerico and
                  Stefano Mintchev and
                  Alessandro Giusti and
                  Boris Gromov and
                  Kamilo Melo and
                  Tomislav Horvat and
                  Cesar Cadena and
                  Marco Hutter and
                  Auke Jan Ijspeert and
                  Dario Floreano and
                  Luca Maria Gambardella and
                  Roland Siegwart and
                  Davide Scaramuzza},
  bibsource    = {dblp computer science bibliography, https://dblp.org},
  doi          = {10.1002/ROB.21887},
  journal      = {J. Field Robotics},
  number       = {7},
  pages        = {1171--1191},
  title        = {{T}he current state and future outlook of rescue robotics},

  volume       = {36},
  year         = {2019}
}

@article{bravo2021internet,
  author       = {Juan Bravo{-}Arrabal and
                  Manuel Toscano{-}Moreno and
                  Juan Jes{\'{u}}s Fern{\'{a}}ndez Lozano and
                  Anthony Mandow and
                  Jos{\'{e}} Antonio G{\'{o}}mez{-}Ruiz and
                  Alfonso Garc{\'{\i}}a{-}Cerezo},
  bibsource    = {dblp computer science bibliography, https://dblp.org},
  doi          = {10.3390/S21237843},
  journal      = {Sensors},
  number       = {23},
  pages        = {7843},
  title        = {{T}he {I}nternet of {C}ooperative {A}gents {A}rchitecture ({X}-{I}o{C}{A}) for {R}obots, {H}ybrid {S}ensor {N}etworks, and {MEC} {C}enters in {C}omplex {E}nvironments: {A} {S}earch and {R}escue {C}ase {S}tudy},

  volume       = {21},
  year         = {2021}
}

@article{planke2020cyber,
  title={A cyber-physical-human system for one-to-many {UAS} operations: Cognitive load analysis},
  author={Planke, Lars J and Lim, Yixiang and Gardi, Alessandro and Sabatini, Roberto and Kistan, Trevor and Ezer, Neta},
  journal={Sensors},
  volume={20},
  number={19},
  pages={5467},
  year={2020},
  publisher={MDPI}
}

@inproceedings{christensen2022herd,
  title={The {HERD} Project: Human-Multi-Robot Interaction in Search \& Rescue and in Farming},
  author={Christensen, Anders Lyhne and Gr{\o}ntved, Kasper Andreas R{\o}mer and Hoang, Maria-Theresa Oanh and van Berkel, Niels and Skov, Mikael and Scovill, Alea and Edwards, Gareth and Geipel, Kenneth Richard and Dalgaard, Lars and Lundquist, Ulrik Pagh Schultz and others},
  booktitle={Adjunct Proceedings of the IEEE/RSJ International Conference on Intelligent Robots and Systems},
  pages={1--4},
  year={2022}
}

@article{jacobsen2023design,
  title={Design of an autonomous cooperative drone swarm for inspections of safety critical infrastructure},
  author={Jacobsen, Rune Hylsberg and Matlekovic, Lea and Shi, Liping and Malle, Nicolaj and Ayoub, Naeem and Hageman, Kaspar and Hansen, Simon and Nyboe, Frederik Falk and Ebeid, Emad},
  journal={Applied Sciences},
  volume={13},
  number={3},
  pages={1256},
  year={2023},
  publisher={MDPI}
}

@article{sapkota2025uavs,
  title={{UAV}s Meet Agentic {AI}: A Multidomain Survey of Autonomous Aerial Intelligence and Agentic {UAV}s},
  author={Sapkota, Ranjan and Roumeliotis, Konstantinos I and Karkee, Manoj},
  journal={arXiv preprint arXiv:2506.08045},
  year={2025}
}

@inproceedings{grontved2023decentralized,
  title={Decentralized multi-{UAV} trajectory task allocation in search and rescue applications},
  author={Gr{\o}ntved, Kasper AR and Lundquist, Ulrik PS and Christensen, Anders Lyhne},
  booktitle={2023 21st International Conference on Advanced Robotics (ICAR)},
  pages={35--41},
  year={2023},
  organization={IEEE}
}

@inproceedings{bahodi2025before,
  title={Before It Falls: Supporting Drone Fleet Management Through Battery Visualizations},
  author={Bahodi, Maria-Theresa and Lau, Nathan and van Berkel, Niels and Gr{\o}ntved, Kasper Andreas R{\o}mer and Skov, Mikael B and Merritt, Timothy},
  booktitle={IFIP Conference on Human-Computer Interaction},
  pages={347--370},
  year={2025},
  organization={Springer}
}

@article{bodker-et-2018,
author = {B\o{}dker, Susanne and Kyng, Morten},
title = {Participatory Design that Matters—Facing the Big Issues},
year = {2018},
issue_date = {February 2018},
publisher = {Association for Computing Machinery},
address = {New York, NY, USA},
volume = {25},
number = {1},
issn = {1073-0516},
url = {https://doi.org/10.1145/3152421},
doi = {10.1145/3152421},
journal = {ACM Trans. Comput.-Hum. Interact.},
month = feb,
articleno = {4},
numpages = {31}
}

@article{endsley_toward_1995,
	title = {Toward a {Theory} of {Situation} {Awareness} in {Dynamic} {Systems}},
	volume = {37},
	issn = {0018-7208},
	url = {https://doi.org/10.1518/001872095779049543},
	doi = {10.1518/001872095779049543},
	number = {1},
	urldate = {2024-01-10},
	journal = {Human Factors},
	publisher = {SAGE Publications Inc},
	author = {Endsley, Mica R.},
	month = mar,
	year = {1995},
	pages = {32--64},
}

@inproceedings{bahodi_show_2024,
	address = {New York, NY, USA},
	series = {{TAS} '24},
	title = {Show {Me} {What}'s {Wrong}: {Impact} of {Explicit} {Alerts} on {Novice} {Supervisors} of a {Multi}-{Robot} {Monitoring} {System}},
	isbn = {979-8-4007-0989-0},
	shorttitle = {Show {Me} {What}'s {Wrong}},
	url = {https://dl.acm.org/doi/10.1145/3686038.3686069},
	doi = {10.1145/3686038.3686069},
	urldate = {2025-09-02},
	booktitle = {Proceedings of the {Second} {International} {Symposium} on {Trustworthy} {Autonomous} {Systems}},
	publisher = {Association for Computing Machinery},
	author = {Bahodi, Maria-Theresa and van Berkel, Niels and Skov, Mikael and Merritt, Timothy},
	month = sep,
	year = {2024},
	pages = {1--17},
}

@inproceedings{kim2020user,
  author = {Kim, Lawrence H. and Drew, Daniel S. and Domova, Veronika and Follmer, Sean},
  title = {User-defined Swarm Robot Control},
  year = {2020},
  isbn = {9781450367080},
  publisher = {Association for Computing Machinery},
  address = {New York, NY, USA},
  doi = {10.1145/3313831.3376814},
  booktitle = {Proceedings of the 2020 CHI Conference on Human Factors in Computing Systems},
  pages = {1--13},
  numpages = {13},
  location = {Honolulu, HI, USA},
  series = {CHI '20}
}

@article{chung2018survey,
  author = {Chung, Soon-Jo and Paranjape, Aditya Avinash and Dames, Philip and Shen, Shaojie and Kumar, Vijay},
  journal = {IEEE Transactions on Robotics},
  title = {A Survey on Aerial Swarm Robotics},
  year = {2018},
  volume = {34},
  number = {4},
  pages = {837--855},
  doi = {10.1109/TRO.2018.2857475}
}

@inproceedings{singh2023progprompt,
  author = {Singh, Ishika and Blukis, Valts and Mousavian, Arsalan and Goyal, Ankit and Xu, Danfei and Tremblay, Jonathan and Fox, Dieter and Thomason, Jesse and Garg, Animesh},
  title = {ProgPrompt: Generating Situated Robot Task Plans using Large Language Models},
  booktitle = {2023 IEEE International Conference on Robotics and Automation (ICRA)},
  year = {2023},
  pages = {11523--11530},
  doi = {10.1109/ICRA48891.2023.10161317},
  organization = {IEEE}
}

@inproceedings{xia2023llm_agents_industry,
  author = {Xia, Yuchen and Shenoy, Manthan and Jazdi, Nasser and Weyrich, Michael},
  title = {Towards autonomous system: flexible modular production system enhanced with large language model agents},
  booktitle = {2023 IEEE International Conference on Emerging Technologies and Factory Automation (ETFA)},
  year = {2023},
  organization = {IEEE},
  doi = {10.1109/ETFA54631.2023.10275362}
}

%
%
%

\end{document}